# Development of Marathi Part of Speech Tagger Using Statistical Approach

Jyoti Singh
Department of Computer Science
Banasthali University
Rajasthan, India
jyoti.singh132@gmail.com

Nisheeth Joshi
Department of Computer Science
Banasthali University
Rajasthan, India
nisheeth.joshi@rediffmail.com

Iti Mathur
Department of Computer Science
Banasthali University
Rajasthan, India
Mathur_iti@rediffmail.com

*Abstract*—Part-of-speech (POS) tagging is a process of assigning the words in a text corresponding to a particular part of speech. A fundamental version of POS tagging is the identification of words as nouns, verbs, adjectives etc. For processing natural languages, Part of Speech tagging is a prominent tool. It is one of the simplest as well as most constant and statistical model for many NLP applications. POS Tagging is an initial stage of linguistics, text analysis like information retrieval, machine translator, text to speech synthesis, information extraction etc. In POS Tagging we assign a Part of Speech tag to each word in a sentence and literature. Various approaches have been proposed to implement POS taggers.
In this paper we present a Marathi part of speech tagger. It is morphologically rich language. Marathi is spoken by the native people of Maharashtra. The general approach used for development of tagger is statistical using Unigram, Bigram, Trigram and HMM Methods. It presents a clear idea about all the algorithms with suitable examples. It also introduces a tag set for Marathi which can be used for tagging Marathi text. In this paper we have shown the development of the tagger as well as compared to check the accuracy of taggers output. The three Marathi POS taggers viz. Unigram, Bigram, Trigram and HMM gives the accuracy of 77.38%, 90.30%, 91.46% and 93.82% respectively.

## I. INTRODUCTION

In this paper we develop a Part of Speech Tagger in Marathi to mark words and punctuation characters in a text with appropriate POS labels for Marathi Text. POS tagging is a very important pre-processing task for Natural language processing activities. Part of Speech tagging for natural language texts are developed using linguistic rule, stochastic models and a combination of both. In this paper we are showing development of simple and efficient automatic taggers for inflectional and derivational language Marathi. Developing POS tagger for Indian languages is difficult job due to morphological richness, lack of peculiar linguistic rules and large annotated corpora. Part-of-speech tagging is a process of assigning the words in a text corresponding to a particular part of speech. Fundamentally Part-of-speech tagging is also called grammatical tagging of text based on both, its definition as well as its context. Parts of speech can be divided into two broad categories: closed classes and open classes. Closed classes are those that have relatively fixed membership. For example, pronouns are categorized in closed class because there is a fixed set of them in English; new pronouns are rarely added. But nouns are in open class because new nouns are continually added in every language.

A Part-Of-Speech Tagger is a piece of software that reads text in some language and assigns parts of speech to each word. There are various approaches of POS tagging, which can be divided into three categories; rule based tagging, statistical tagging and hybrid tagging. The rule based POS tagging model applies a set of hand written rules and uses contextual information to assign POS tags to words. The main drawback of rule based system is that it fails when the text is unknown. The rule based system cannot predict the appropriate tag. Hence for achieving higher accuracy in this system we need to have an exhaustive set of hand coded rules. A statistical approach includes frequency and probability. The simplest statistical approach finds out the most frequently used tag for a specific word from the annotated training data and uses this information to tag that word in the unannotated text. The problem with this approach is that it can come up with sequences of tags for sentences that are not acceptable according to the grammar rules of a language. There is another approach which is the hybrid one. This may perform better than statistical or rule based approaches. The hybrid approach first uses the probabilistic features of the statistical method and then applies the set of hand coded language rules. This paper discuss the different types of statistical tagging approaches which are Unigram, Bigram, Trigram and HMM, also shows the evaluation done and the comparative study of their result.

## II. PROBLEMS OF PART OF SPEECH TAGGING

The main problem in part of speech tagging is ambiguous words. There may be many words which can have more than one tag. Sometimes it happens that a word has same POS but has different meaning in different context. To solve this problem we consider the context instead of taking single word. For example-

1- रंगून/**NNP** ला/PSP गाण्याच्या/JJ कार्यक्रमात/NN श्याम/ NNP रंगून/**RB** गेला/VM ./SYM

The same word 'रंगून' is given a different label in a same sentence. In the first case it is termed as a proper noun. In the second case it is termed as an adjective as it is referring to the feeling of any person. Since first word 'रंगून' occurs in a sentence as subject which is followed by a postposition, that is why it is labeled as NNP. Whereas in second time 'रंगून' comes between a main verb and a noun so it is assigned as an adverb. POS Tagging tries to correctly identify a POS of a word by looking at the context (surrounding words) in a sentence.

2- वंदना/ **NNP** नी/ PSP देवीची/ **NNP** वंदना/ VM केली/ VAUX ./SYM

Like above example here, same word 'वंदना' is given a different label in a same sentence. In the first case it is termed as a proper noun. In the second case it is termed as a main verb as it is referring to any work done. Since first word 'वंदना' occur in a sentence as subject and after that there is a postposition therefore it is labeled as NNP and in second time 'वंदना' comes before a helping verb and after a noun so it is assigned as main verb.

III. PREVIOUS WORK ON INDIAN LANGUAGE POS TAGGING

Different approaches have been used for POS tagging and enormous research works have been done in this area Hninn Myint et. al. [2] proposed a Bigram Part of Speech tagger for Myanmar, they developed a bigram POS tagger using Baum Welch and Viterbi algorithm for tagging and decoding purpose respectively and they achieved 90% accuracy. The statistical approaches [4, 5, 6, 9] use tagset to develop the tagger and for finding most probable tags, they used training corpus. All the statistical methods cited above are generally based on Unigram, Bigram, Trigram and HMM and shows the accuracy of 92.13%, 85.56%, 91.23% and 95.64% for Indian languages.

The most notable section in the area of POS tagging is work done using CRF and SVM [3, 7, 10] proposed a Manipuri, Tamil and Gujarati POS tagger respectively. Their taggers show machine learning algorithms and for that work they have applied CRF and SVM. Singh et. al. [8] in 2008 proposed Part-of-Speech Tagging for Grammar Checking of Punjabi. In this paper, they have discussed the issues concerning the development of a POS tagset and a POS tagger for the use as a part of the project on developing an automated grammar checking system for Punjabi Language. They reported an accuracy of 80.29% for their tagger.

Reddy and Sharoff [11] proposed Cross Language POS Tagger (and other Tools) for Indian Languages: An Experiment with Kannada using Telugu Resources, they have used TnT (Brants, 2000), a popular implementation of the second-order Markov model for POS tagging. Kumar et. Al [12] presented Part of Speech Tagger for Morphologically rich Indian Languages: A survey. In this paper they have reported about different POS taggers based on different languages and methods. Kumar et. al. [13] presented Building Feature Rich POS Tagger for Morphologically Rich Languages: Experiences in Hindi. The tagger is based on maximum entropy Markov model with a rich set of features capturing the lexical and morphological characteristics of the language. This system achieved the best accuracy of 94.89% and an average accuracy of 94.38%.

IV. POS TAGSET

Depending on some general principle of tagset design strategy, a number of POS tagsets have been developed by different organizations. For POS annotation texts in Marathi, we have used tagset developed by IIIT Hyderabad (Bharti, et. al., 2006) [1]. They have around 20 relations (semantic tags) and 15 node level tags or syntactic tags. Subsequently, a common tagset has been designed for POS tagging and chunking of a large group of the Indian languages. The tagset consist of 26 lexical tags. The tagset was designed based on the lexical category of a word.

TABLE I
POS TAGSET FOR MARATHI

| S.No. | Tag | Description (Tag Used for) | Example |
|---|---|---|---|
| 1. | NN | Common Nouns | मुलगा, साखर, मंडळी, सैन्य, चांगुलपणा |
| 2. | NST | Noun Denoting Spatial and Temporal Expressions | मागे, पुढे, वर, खाली |
| 3. | NNP | Proper Nouns (name of person) | मोहन, राम, सुरेश |
| 4. | PRP | Pronoun | मी,आम्ही,तुम्ही |
| 5. | DEM | Demonstrative | तो, ती, ते, हा, ही |
| 6. | VM | Verb Main (Finite or Non-Finite) | बसणे, दिसणे, लिहिणे |

| S.No. | Tag | Description (Tag Used for) | Example |
|---|---|---|---|
| 7. | VAUX | Verb Auxiliary | नाही, नको, करणे ,हवे, नये |
| 8. | JJ | Adjective (Modifier of Noun) | उत्साही, श्रेष्ठ,बळवान |
| 9. | RB | Adverb (Modifier of Verb) | आता, काल, कधी, नेहमी |
| 10. | PSP | Postposition | आणि, वर, कडे |
| 11. | RP | Particles | भी, तो, ही |
| 12. | QF | Quantifiers | बहुत, थोडा, कम |
| 13. | QC | Cardinals | एक, दोन, तीन |
| 14. | CC | Conjuncts (Coordinating and Subordinating) | आणि,केव्हा,तेव्हां, जर, तर |
| 15. | WQ | Question Words | काय, कधी, कुठे |
| 16. | QO | Ordinals | पहिला,दुसरा, तिसरा |
| 17. | INTF | Intensifier | खूप, फार, पुष्कळ |
| 18. | INJ | Interjection | आहा, छान, अगो, हाय |
| 19. | NEG | Negative | नाही,नको |
| 20. | SYM | Symbol | ? , ; : ! |
| 21. | XC | Compounds | काळेमांजर-काळमांजर, तेलपाणी-तेलवणी |
| 22. | RDP | Reduplications | जवळ-जवळ |
| 23. | UNK | Foreign Words | English, ગુજરાતી |

V. METHEDOLOGY

In our work, we have Marathi corpus. We are using statistical approach for POS tagging i.e. We train and test our model for this we have to calculate frequency and probability of words of given corpus. To train our system we used 7000 sentences (1, 95,647) words from tourism domain.

**(i) Unigram**
A POS Tagger Based on Unigram model assigns each word to its most common tag. In this model, we only consider one word at a time. Generally unigram method for calculating part of speech are based on simple statistical model i.e. basic idea behind that, is calculation of unigram probability. For this reason, unigram tagger is also called 1-gram tagger. In this method for each word, assigns the tag that is most likely for that particular word. Figure1 shows this phenomenon. Annotated data can also be used to train the corpus.
For calculating the unigram probability, we first determined that how many times each word occurs in the corpus. So the equation (1) shows the phenomenon-

$$P(t_i/w_i) = freq(w_i/t_i)/freq(w_i) \ldots\ldots\ldots (1)$$

Here Probability of tag given word is computed by frequency count of word given tag divided by frequency count of that particular word. Corresponding probabilities will be checked after calculating frequency. At last on the basis of those probabilities final tagged output will be generated.

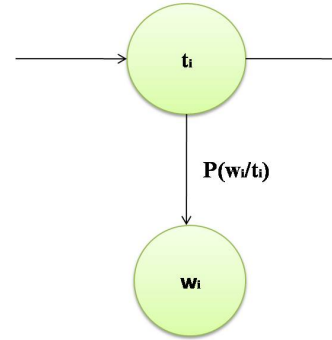

Fig. 1 Working of Unigram Model

**(ii) Bigram**
This section presents a system based on bigram method. The basic idea behind all the statistical method is to capture most likely tag sequences for text. Bigram tagger makes tag suggestion based on preceding tag i.e. it take two tags: the preceding tag and current tag into account. Unlike Unigram tagger it considers the context when assigning a tag to the current word. Bigram tagger assumes that probability of tags depend on previous tags. So this phenomenon can show by equation (2)-

$$P(t_i/w_i) = P(w_i/t_i) \cdot P(t_i/t_{i-1}) \ldots\ldots\ldots (2)$$

Here P ($w_i/t_i$) is the probability of current word given current tag
P ($t_i/t_{i-1}$) is the probability of a current tag given the previous tag
These probabilities are computed by equation (3)

$$P(t_i/t_{i-1}) = f(t_{i-1}, t_i)/f(t_{i-1}) \ldots\ldots\ldots\ldots (3)$$

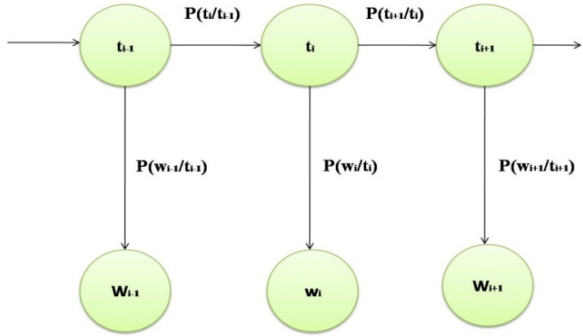

Fig. 2 Working of Bigram Model

**(iii) Trigram**
For describing Trigram Model for POS tagger, our main aim is to perform POS Tagging to determine the most likely tag for a word, given the previous two tags. Working diagram of trigram model is described in figure 3. For trigrams, the probability of a sequence is just the product of conditional probabilities of its trigrams. So if $t_1, t_2 \ldots t_n$ are tag sequence and $w_1, w_2 \ldots w_n$ are corresponding word sequence then the equation (4) explains this fact-

$$P(t_i/w_i) = P(w_i/t_i) \cdot P(t_i/t_{i-2}, t_{i-1}) \ldots\ldots (4)$$

Where $t_i$ denotes tag sequence and $w_i$ denote word sequence.
P ($w_i/t_i$) is the probability of current word given current tag.
Here, $P(t_i|t_{i-2}t_{i-1})$ is the probability of a current tag given the previous two tags.

This provides the transition between the tags and helps to capture the context of the sentence. These probabilities are computed by equation (5).

$$P(t_i/t_{i-2}, t_{i-1}) = f(t_{i-2}, t_{i-1}, t_i)/f(t_{i-2}, t_{i-1}) \ldots\ldots (5)$$

Each tag transition probability is computed by calculating the frequency count of two tags which come together in the corpus divided by the frequency count of the previous two tags coming in the corpus.

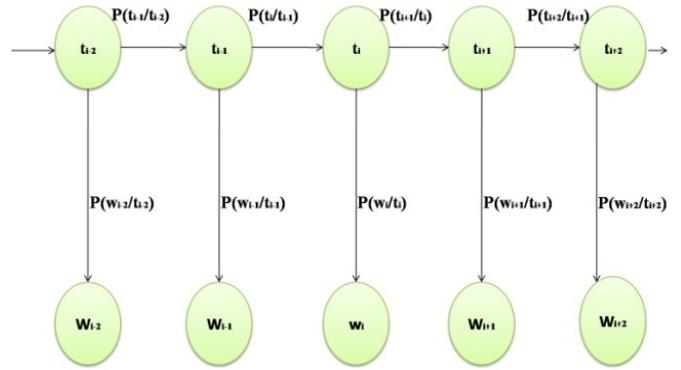

Fig. 3 Working of Trigram Model

**(iv) HMM**
A HMM is Statistical Model which can be used to generate tag sequences. Basic idea of HMM is to determine the most likely tag sequences. For this purpose we have to calculate Transition probability. Transition probability shows the probability of travelling between two tags i.e. forward tag and backward tags.
The Transition probability is generally estimated based on previous tags and future tags with the sequence provided as an input. The following equation (6) explains this idea-

$$P(t_i/w_i) = P(t_i/t_{i-1}) \cdot P(t_{i+1}/t_i) \cdot P(w_i/t_i) \ldots\ldots (6)$$

P ($t_i/t_{i-1}$) is the probability of current tag given previous tag.
P ($t_{i+1}/t_i$) is the probability of future tag given current tag.
P ($w_i/t_i$) Probability of word given current tag
It is calculated as-

$$P(w_i/t_i) = freq(t_i, w_i)/freq(t_i) \ldots\ldots\ldots (7)$$

This is done because we know that it is more likely for some tags to precede the other tags.
In HMM we consider the context of tags with respect to the current tag. It assigns the best tag to a word by calculating the forward and backward probabilities of tags along with the sequence provided as an input. Powerful feature of HMM is context description which can decides the tag for a word by looking at the tag of the previous word and the tag of the future word. Figure 4 shows the idea behind the model.

## VI. EVALUATION

We apply Unigram, Bigram, Trigram and HMM methods on Marathi text. In order to measure the performance of our systems, we developed a test corpus of 1000 sentences (25744 words). We finally report results of all POS taggers in terms of accuracy.

The accuracy was calculated by using this formula:

**Accuracy (%) = (No. of correctly tagged token/ Total no. of POS tags in the text)*100**

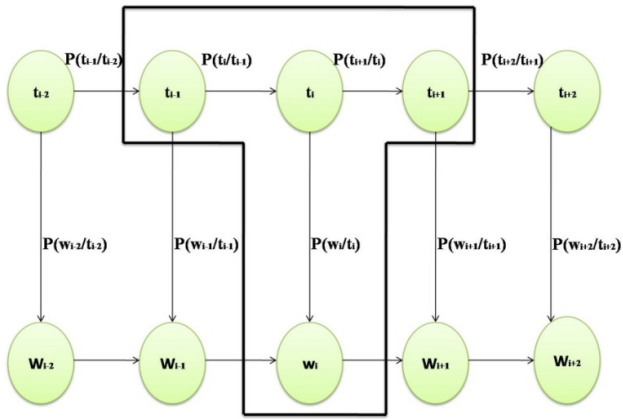

Fig. 4 Working of HMM Model

Test scores of our system are as follows:

**For Unigram:**

For example:
एक/QC हांडी/NN साखर/**NN** भाताला/NN पाव/NN शेर/NN साखर/NN लागते/**VUX**./SYM

In the above example Unigram tagger assigns noun to word 'साखर' and auxiliary verb to 'लागते'. But ideally we know that 'साखर' is an adjective and 'लागते' is main verb.

No. of Correct POS tags assigned by the system = 19921
No. of POS tag in the text = 25744
Thus the accuracy of the system is 77.39%.

**For Bigram:**
एक/QC हांडी/NN साखर/JJ भाताला/**NNP** पाव/NN शेर/NN साखर/NN लागते/VM./SYM

In the above example Bigram tagger assigns proper noun to word 'भाताला', which is wrong assessment by tagger.

No. of Correct POS tags assigned by the system = 23249
No. of POS tag in the text = 25744
Thus the accuracy of the system is 90.30%.

**For Trigram:**
एक/QC हांडी/NN साखर/JJ भाताला/NN पाव/NN शेर/NN साखर/NN लागते/VM. /SYM

Here Trigram tagger assigns correct tag to each word.

No. of Correct POS tags assigned by the system = 23546
No. of POS tag in the text = 25744
Thus the accuracy of the system is 91.46%.

**For HMM:**
एक/QC हांडी/NN साखर/JJ भाताला/NN पाव/NN शेर/NN साखर/NN लागते/VM./SYM

In above sentence HMM assigns correct tag.

No. of Correct POS tags assigned by the system = 24156
No. of POS tag in the text = 25744
Thus the accuracy of the system is 93.82%.

VII. COMPARISION WITH EXISTING SYSTEMS

Our system results for Part of Speech tagger for Marathi. Some of the systems that are to some extent closes to our system in terms of applied model i.e. HMM and accuracy received are given here for comparison.
A POS tagger for Bangla reports 85.56% accuracy. A system for Hindi reports 92.13% accuracy. Another system for Hindi repots 89.34% accuracy. A model for Malayalam provides accuracy of 90%. Assamese POS Tagger gives 87% of accuracy. Our part of speech tagger gives 93.82% accuracy which seems better.

VIII. CONCLUSION

The Part-of-speech tagging is playing an important role in various speech and language processing applications. Currently many tools are available to do this task of part of speech tagging. The POS taggers described here is very simple and efficient for automatic tagging, but the morphological complexity of the Marathi makes it little hard. The results of all the taggers are impressive. The performance of the current system is good and the results achieved by methods are excellent. We believe that future enhancements of this work would be to improve the tagging accuracy by increasing the size of tagged corpus.